# Perseus: Randomized Point-based Value Iteration for POMDPs

**Matthijs T. J. Spaan**  MTJSPAAN@SCIENCE.UVA.NL
**Nikos Vlassis**  VLASSIS@SCIENCE.UVA.NL
*Informatics Institute, University of Amsterdam*
*Kruislaan 403, 1098 SJ Amsterdam, The Netherlands*

## Abstract

Partially observable Markov decision processes (POMDPs) form an attractive and principled framework for agent planning under uncertainty. Point-based approximate techniques for POMDPs compute a policy based on a finite set of points collected in advance from the agent's belief space. We present a randomized point-based value iteration algorithm called PERSEUS. The algorithm performs approximate value backup stages, ensuring that in each backup stage the value of each point in the belief set is improved; the key observation is that a single backup may improve the value of many belief points. Contrary to other point-based methods, PERSEUS backs up only a (randomly selected) subset of points in the belief set, sufficient for improving the value of each belief point in the set. We show how the same idea can be extended to dealing with continuous action spaces. Experimental results show the potential of PERSEUS in large scale POMDP problems.

## 1. Introduction

A major goal of Artificial Intelligence is to build intelligent agents (Russell & Norvig, 2003). An intelligent agent, whether physical or simulated, should be able to autonomously perform a given task, and is often characterized by its sense–think–act loop: it uses sensors to observe the environment, considers this information to decide what to do, and executes the chosen action. The agent influences its environment by acting and can detect the effect of its actions by sensing: the environment closes the loop. In this work we are interested in computing a *plan* that maps sensory input to the optimal action to execute for a given task. We consider types of domains in which an agent is uncertain about the exact consequence of its actions. Furthermore, it cannot determine with full certainty the state of the environment with a single sensor reading, i.e., the environment is only partially observable to the agent.

Planning under these kinds of uncertainty is a challenging problem as it requires reasoning over all possible futures given all possible histories. Partially observable Markov decision processes (POMDPs) provide a rich mathematical framework for acting optimally in such partially observable and stochastic environments (Dynkin, 1965; Åström, 1965; Aoki, 1965; Sondik, 1971; Lovejoy, 1991; Kaelbling, Littman, & Cassandra, 1998). The POMDP defines a sensor model specifying the probability of observing a particular sensor reading in a specific state and a stochastic transition model which captures the uncertain outcome of executing an action. The agent's task is defined by the reward it receives at each time step and its goal is to maximize the discounted cumulative reward. Assuming discrete models, the POMDP framework allows for capturing all uncertainty introduced by the transition and observation model by defining and operating on the *belief state* of an agent. A belief state is a probability distribution over all states and summarizes all information regarding the past.

The use of belief states allows one to transform the original discrete state POMDP into a continuous state Markov decision process (MDP), which in turn can be solved by corresponding MDP techniques (Bertsekas & Tsitsiklis, 1996). However, the optimal value function in a POMDP exhibits particular structure (it is piecewise linear and convex) that one can exploit in order to facilitate the solving. Value iteration, for instance, is a method for solving POMDPs that builds a sequence of value function estimates which converge to the optimal value function for the current task (Sondik, 1971). The value function is parameterized by a finite number of hyperplanes, or vectors, over the belief space, which partition the belief space in a finite amount of regions. Each vector maximizes the value function in a certain region and has an action associated with it, which is the optimal action to take for beliefs in its region. Computing the next value function estimate—looking one step deeper into the future—requires taking into account all possible actions the agent can take and all subsequent observations it may receive. Unfortunately, this leads to an exponential growth of vectors with the planning horizon. Many of the computed vectors will be useless in the sense that their maximizing region is empty, but identifying and subsequently pruning them is an expensive operation.

Exact value iteration algorithms (Sondik, 1971; Cheng, 1988; Kaelbling et al., 1998) search in each value iteration step the complete belief simplex for a minimal set of belief points that generate the necessary set of vectors for the next horizon value function. This typically requires linear programming and is therefore costly in high dimensions. Zhang and Zhang (2001) argued that value iteration still converges to the optimal value function if exact value iteration steps are interleaved with approximate value iteration steps in which the new value function is an upper bound to the previously computed value function. This results in a speedup of the total algorithm, however, linear programming is again needed in order to ensure that the new value function is an upper bound to the previous one over the complete belief simplex. In general, computing exact solutions for POMDPs is an intractable problem (Papadimitriou & Tsitsiklis, 1987; Madani, Hanks, & Condon, 1999), calling for approximate solution techniques (Lovejoy, 1991; Hauskrecht, 2000).

In practical tasks one would like to compute solutions only for those parts of the belief simplex that are reachable, i.e., that can be actually encountered by interacting with the environment. This has recently motivated the use of approximate solution techniques which focus on the use of a sampled set of *belief points* on which planning is performed (Hauskrecht, 2000; Poon, 2001; Roy & Gordon, 2003; Pineau, Gordon, & Thrun, 2003; Spaan & Vlassis, 2004), a possibility already mentioned by Lovejoy (1991). The idea is that instead of planning over the complete belief space of the agent (which is intractable for large state spaces), planning is carried out only on a limited set of prototype beliefs that have been sampled by letting the agent interact (randomly) with the environment. PBVI (Pineau et al., 2003), for instance, builds successive estimates of the value function by updating the value and its gradient only at the points of a (dynamically growing) belief set.

In this work we describe Perseus, a randomized point-based value iteration algorithm for POMDPs (Vlassis & Spaan, 2004; Spaan & Vlassis, 2004). Perseus operates on a large set of beliefs which are gathered by simulating random interactions of the agent with the POMDP environment. On this belief set a number of value backup stages are performed.

The algorithm ensures that in each backup stage the value of each point in the belief set is improved (or at least does not decrease). Contrary to other point-based methods, Perseus backs up only a random subset of belief points; the key observation is that a single backup may improve the value of many points in the set. This allows us to compute value functions that consist of only a small number of vectors (relative to the belief set size), leading to significant speedups. We evaluate the performance of Perseus on benchmark problems from literature, and show that it is very competitive to other methods in terms of solution quality and computation time.

Furthermore, we extend Perseus to compute plans for agents which have a continuous (or very large discrete) set of actions at their disposal (Spaan & Vlassis, 2005). Examples include navigating to an arbitrary location, or rotating a pan-and-tilt camera at any desired angle. Most work on POMDP solution techniques targets discrete action spaces; exceptions include the application of a particle filter to a continuous state and action space (Thrun, 2000) and certain policy search methods (Ng & Jordan, 2000; Baxter & Bartlett, 2001). We report on experiments in a domain in which an agent equipped with proximity sensors can move at a continuous heading and distance, and we present experimental results from a navigation task involving a mobile robot with omnidirectional vision in a perceptually aliased office environment.

The remainder of the paper is structured as follows: in Section 2 we review the POMDP framework from an AI perspective, and we discuss exact methods for solving POMDPs and their tractability problems. Next, we outline a class of approximate value iteration algorithms, the so-called point-based techniques. In Section 3 we describe and discuss the Perseus algorithm, as well as the extension to continuous action spaces. Related work on approximate techniques for POMDP planning is discussed in Section 4. We present experimental results from several problem domains in Section 5. Finally, we wrap up with some conclusions in Section 6.

## 2. Partially Observable Markov Decision Processes

A partially observable Markov decision process (POMDP) models the repeated interaction of an agent with a stochastic environment, parts of which are hidden from the agent's view. The agent's goal is to perform a task by choosing actions which fulfill the task best. Stated otherwise, the agent has to compute a plan that optimizes the given performance measure. We assume that time is discretized in time steps of equal length, and at the start of each step the agent has to execute an action. At each time step the agent also receives a scalar reward from the environment, and the performance measure directs the agent to maximize the cumulative reward it can gather. The reward signal allows one to define a task for the agent, e.g., one can give the agent a large positive reward when it accomplishes a certain goal and a small negative reward for each action leading up to it. In this way the agent is steered toward finding the plan which will let it accomplish its goal as fast as possible.

The POMDP framework models stochastic environments in which an agent is uncertain about the exact effect of executing a certain action. This uncertainty is captured by a probabilistic transition model as is the case in a fully observable Markov decision process (MDP) (Sutton & Barto, 1998; Bertsekas & Tsitsiklis, 1996). An MDP defines a transition model which specifies the probabilistic effect of how each action changes the state. Extending the MDP setting, a POMDP also deals with uncertainty resulting from the agent's imperfect sensors. It allows for planning in environments which are only partially observable to the agent, i.e., environments in which the agent cannot determine with full certainty the true state of the environment. In general the partial observability stems from two sources: (1) multiple states give the same sensor reading, in case the agent can only sense a limited part of the environment, and (2) its sensor readings are noisy: observing the same state can result in different sensor readings. The partial observability can lead to "perceptual aliasing": different parts of the environment appear similar to the agent's sensor system, but require different actions. The POMDP captures the partial observability by a probabilistic observation model, which relates possible observations to states.

More formally, a POMDP assumes that at any time step the environment is in a state $s \in S$, the agent takes an action $a \in A$ and receives a reward $r(s, a)$ from the environment as a result of this action, while the environment switches to a new state $s'$ according to a known stochastic transition model $p(s'|s, a)$. The Markov property entails that $s'$ only depends on the previous state $s$ and the action $a$. The agent then perceives an observation $o \in O$, that may be conditional on its action, which provides information about the state $s'$ through a known stochastic observation model $p(o|s, a)$. All sets $S$, $O$, and $A$ are assumed discrete and finite here (but we will generalize to continuous $A$ in Section 3.3).

In order for an agent to choose its actions successfully in partially observable environments some form of memory is needed, as the observations the agent receives do not provide an unique identification of $s$. Given the transition and observation model the POMDP can be transformed to a belief-state MDP: the agent summarizes all information about its past using a belief vector $b(s)$. The belief $b$ is a probability distribution over $S$, which forms a Markovian signal for the planning task. All beliefs are contained in a $(|S| - 1)$-dimensional simplex $\Delta$, which means we can represent a belief using $|S| - 1$ numbers. Each POMDP problem assumes an initial belief $b_0$, which for instance can be set to a uniform distribution over all states (representing complete ignorance regarding the initial state of the environment). Every time the agent takes an action $a$ and observes $o$, its belief is updated by Bayes' rule:

$$b_a^o(s') = \frac{p(o|s', a)}{p(o|a, b)} \sum_{s \in S} p(s'|s, a) b(s), \qquad (1)$$

where $p(o|a, b) = \sum_{s' \in S} p(o|s', a) \sum_{s \in S} p(s'|s, a) b(s)$ is a normalizing constant.

As we discussed above, the goal of the agent is to choose actions which fulfill its task as well as possible, i.e., to compute an optimal plan. Such a plan is called a policy $\pi(b)$ and maps beliefs to actions. Note that, contrary to MDPs, the policy $\pi(b)$ is a function over a continuous set of probability distributions over $S$. A policy $\pi$ can be characterized by a value function $V^\pi : \Delta \to \mathbb{R}$ which is defined as the expected future discounted reward $V^\pi(b)$ the agent can gather by following $\pi$ starting from belief $b$:

$$V^\pi(b) = E_\pi \Big[ \sum_{t=0}^{\infty} \gamma^t r(b_t, \pi(b_t)) \Big| b_0 = b \Big], \qquad (2)$$

where $r(b_t, \pi(b_t)) = \sum_{s \in S} r(s, \pi(b_t)) b_t(s)$, and $\gamma$ is a discount rate, $0 \leq \gamma < 1$. The discount rate ensures a finite sum and is usually chosen close to 1. A policy $\pi$ which maximizes $V^\pi$ is called an optimal policy $\pi^*$; it specifies for each $b$ the optimal action to execute at the current step, assuming the agent will also act optimally at future time steps. The value of an optimal policy $\pi^*$ is defined by the optimal value function $V^*$, that satisfies the Bellman optimality equation $V^* = HV^*$:

$$V^*(b) = \max_{a \in A} \left[ \sum_{s \in S} r(s,a) b(s) + \gamma \sum_{o \in O} p(o|a,b) V^*(b_a^o) \right], \tag{3}$$

with $b_a^o$ given by (1), and $H$ is the Bellman backup operator (Bellman, 1957). When (3) holds for every $b \in \Delta$ we are ensured the solution is optimal.

$V^*$ can be approximated by iterating a number of stages, as we will see in the next section, at each stage considering a step further into the future. For problems with a finite planning horizon $V^*$ will be piecewise linear and convex (PWLC) (Smallwood & Sondik, 1973), and for infinite horizon tasks $V^*$ can be approximated arbitrary well by a PWLC value function. We parameterize a value function $V_n$ at stage $n$ by a finite set of vectors (hyperplanes) $\{\alpha_n^i\}$, $i = 1, \ldots, |V_n|$. Additionally, with each vector an action $a(\alpha_n^i) \in A$ is associated, which is the optimal one to take in the current step. Each vector defines a region in the belief space for which this vector is the maximizing element of $V_n$. These regions form a partition of the belief space, induced by the piecewise linearity of the value function. Examples of a value function for a two state POMDP are shown in Fig. 1(a) and 1(d). Given a set of vectors $\{\alpha_n^i\}_{i=1}^{|V_n|}$ at stage $n$, the value of a belief $b$ is given by

$$V_n(b) = \max_{\{\alpha_n^i\}_i} b \cdot \alpha_n^i, \tag{4}$$

where $(\cdot)$ denotes inner product. The gradient of the value function at $b$ is given by the vector $\alpha_n^b = \arg\max_{\{\alpha_n^i\}_i} b \cdot \alpha_n^i$, and the policy at $b$ is given by $\pi(b) = a(\alpha_n^b)$.

## 2.1 Exact Value Iteration

Computing an optimal plan for an agent means solving the POMDP, and a classical method is value iteration (Puterman, 1994). In the POMDP framework, value iteration involves approximating $V^*$ by applying the exact dynamic programming operator $H$ above, or some approximate operator $\tilde{H}$, to an initially piecewise linear and convex value function $V_0$. For $H$, and for many commonly used $\tilde{H}$, the produced intermediate estimates $V_1, V_2, \ldots$ will also be piecewise linear and convex. The main idea behind many value iteration algorithms for POMDPs is that for a given value function $V_n$ and a particular belief point $b$ we can easily compute the vector $\alpha_{n+1}^b$ of $HV_n$ such that

$$\alpha_{n+1}^b = \arg\max_{\{\alpha_{n+1}^i\}_i} b \cdot \alpha_{n+1}^i, \tag{5}$$

where $\{\alpha_{n+1}^i\}_{i=1}^{|HV_n|}$ is the (unknown) set of vectors for $HV_n$. We will denote this operation $\alpha_{n+1}^b = \text{backup}(b)$. It computes the optimal vector for a given belief $b$ by back-projecting all vectors in the current horizon value function one step from the future and returning the vector that maximizes the value of $b$. In particular, defining $r_a(s) = r(s,a)$ and using (1),

(3), and (4) we have:

$$V_{n+1}(b) = \max_a \left[ b \cdot r_a + \gamma \sum_o p(o|a,b) V_n(b_a^o) \right] \qquad (6)$$

$$= \max_a \left[ b \cdot r_a + \gamma \sum_o p(o|a,b) \max_{\{\alpha_n^i\}_i} \sum_{s'} b_a^o(s') \alpha_n^i(s') \right] \qquad (7)$$

$$= \max_a \left[ b \cdot r_a + \gamma \sum_o \max_{\{\alpha_n^i\}_i} \sum_{s'} p(o|s',a) \sum_s p(s'|s,a) b(s) \alpha_n^i(s') \right] \qquad (8)$$

$$= \max_a \left[ b \cdot r_a + \gamma \sum_o \max_{\{g_{a,o}^i\}_i} b \cdot g_{a,o}^i \right], \qquad (9)$$

where

$$g_{a,o}^i(s) = \sum_{s'} p(o|s',a) p(s'|s,a) \alpha_n^i(s'). \qquad (10)$$

Applying the identity $\max_j b \cdot \alpha_j = b \cdot \arg\max_j b \cdot \alpha_j$ in (9) twice, we can compute the vector `backup(b)` as follows:

$$\texttt{backup}(b) = \arg\max_{\{g_a^b\}_{a \in A}} b \cdot g_a^b, \quad \text{where} \qquad (11)$$

$$g_a^b = r_a + \gamma \sum_o \arg\max_{\{g_{a,o}^i\}_i} b \cdot g_{a,o}^i. \qquad (12)$$

Although computing the vector `backup(b)` for a given $b$ is straightforward, locating the (minimal) set of points $b$ required to compute *all* vectors $\cup_b \texttt{backup}(b)$ of $HV_n$ is very costly. As each $b$ has a region in the belief space in which its $\alpha_n^b$ is maximal, a family of algorithms tries to identify these regions (Sondik, 1971; Cheng, 1988; Kaelbling et al., 1998). The corresponding $b$ of each region is called a "witness" point, as it testifies to the existence of its region. Another set of exact POMDP value iteration algorithms do not focus on searching in the belief space, but instead consider enumerating all possible vectors of $HV_n$ and then pruning useless vectors (Monahan, 1982; Cassandra, Littman, & Zhang, 1997).

As an example of exact value iteration let us consider the most straightforward way of computing $HV_n$ due to Monahan (1982). This involves calculating all possible ways $HV_n$ could be constructed, exploiting the known structure of the value function. We operate independent of a particular $b$ now so (12) can no longer be applied. Instead we have to include all ways of selecting $g_{a,o}^i$ for all $o$:

$$HV_n = \bigcup_a G_a, \quad \text{with} \quad G_a = \bigoplus_o \left\{ \frac{1}{|O|} r_a + \gamma g_{a,o}^i \right\}_i, \qquad (13)$$

where $\bigoplus$ denotes the cross-sum operator.[1] Unfortunately, at each stage a number of vectors exponential in $|O|$ are generated: $|A||V_n|^{|O|}$. The regions of many of the generated vectors will be empty and these vectors are useless as such, but identifying and subsequently pruning them requires linear programming which introduces considerable additional cost (e.g., when the state space is large).

---

1. Cross-sum of sets is defined as: $\bigoplus_k R_k = R_1 \oplus R_2 \oplus \ldots \oplus R_k$, with $P \oplus Q = \{ p + q \mid p \in P,\ q \in Q \}$.

Zhang and Zhang (2001) proposed an alternative approach to exact value iteration, designed to speed up each exact value iteration step. It turns out that value iteration still converges to the optimal value function if exact value update steps are interleaved with approximate update steps in which a new value function $V_{n+1}$ is computed from $V_n$ such that

$$V_n(b) \leq V_{n+1}(b) \leq HV_n(b), \qquad \text{for all } b \in \Delta. \tag{14}$$

This additionally requires that the value function is appropriately initialized, by choosing $V_0$ to be a single vector with all its components equal to $\frac{1}{1-\gamma} \min_{s,a} r(s,a)$. Such a vector represents the minimum of cumulative discounted reward obtainable in the POMDP, and is guaranteed to be below $V^*$. Zhang and Zhang (2001) compute $V_{n+1}$ by backing up witness points of $V_n$ for a number of steps. As we saw above, backing up a set of belief points is a relatively cheap operation. Thus, given $V_n$, a number of vectors of $HV_n$ are created by applying `backup` to the witness points of $V_n$, and then a set of linear programs are solved to ensure that $V_{n+1}(b) \geq V_n(b)$, $\forall b \in \Delta$. This is repeated for a number of steps, before an exact value update step takes place. The authors demonstrate experimentally that a combination of approximate and exact backup steps can speed up exact value iteration.

In general, however, computing optimal planning solutions for POMDPs is an intractable problem for any reasonably sized task (Papadimitriou & Tsitsiklis, 1987; Madani et al., 1999). This calls for approximate solution techniques. We will describe next a recent line of research on approximate POMDP algorithms which focus on planning on a fixed set of belief points.

### 2.2 Approximate Value Iteration

The major cause of intractability of exact POMDP solution methods is their aim of computing the optimal action for every possible belief point in $\Delta$. For instance, if we use (13) we end up with a series of value functions whose size grows exponentially in the planning horizon. A natural way to sidestep this intractability is to settle for computing an approximate solution by considering only a finite set of belief points. The backup stage reduces to applying (11) a fixed number of times, resulting in a small number of vectors (bounded by the size of the belief set). The motivation for using approximate methods is their ability to compute successful policies for much larger problems, which compensates for the loss of optimality.

Such approximate POMDP value iteration methods operating on a fixed set of points are explored by Lovejoy (1991) and in subsequent works (Hauskrecht, 2000; Poon, 2001; Pineau et al., 2003; Spaan & Vlassis, 2004). Pineau et al. (2003) for instance, use an approximate backup operator $\tilde{H}_{\text{PBVI}}$ instead of $H$, that computes in each value backup stage the set

$$\tilde{H}_{\text{PBVI}} V_n = \bigcup_{b \in B} \texttt{backup}(b) \tag{15}$$

using a fixed set of belief points $B$. The general assumption underlying these so-called *point-based* methods is that by updating not only the value but also its gradient (the $\alpha$ vector) at each $b \in B$, the resulting policy will generalize well and be effective for beliefs outside the set $B$. Whether or not this assumption is realistic depends on the POMDP's structure and the contents of $B$, but the intuition is that in many problems the set of 'reachable' beliefs (reachable by following an arbitrary policy starting from $b_0$) forms a low dimensional manifold in the belief simplex, and thus can be covered densely enough by a relatively small number of belief points.

Crucial to the control quality of the computed approximate solution is the makeup of $B$. A number of schemes to build $B$ have been proposed. For instance, one could use a regular grid on the belief simplex, computed, e.g., by Freudenthal triangulation (Lovejoy, 1991). Other options include taking all extreme points of the belief simplex or use a random grid (Hauskrecht, 2000; Poon, 2001). An alternative scheme is to include belief points that can be encountered by simulating the POMDP: we can generate trajectories through the belief space by sampling random actions and observations at each time step (Lovejoy, 1991; Hauskrecht, 2000; Poon, 2001; Pineau et al., 2003; Spaan & Vlassis, 2004). This sampling scheme focuses the contents of $B$ to be beliefs that can actually be encountered while experiencing the POMDP model.

The PBVI algorithm (Pineau et al., 2003) is an instance of such a point-based POMDP algorithm. PBVI starts by selecting a small set of beliefs $B_0$, performs a number of backup stages (15) on $B_0$, expands $B_0$ to $B_1$ by sampling more beliefs, performs again a series of backups, and repeats this process until a satisfactory solution has been found (or the allowed computation time expires). The set $B_{t+1}$ grows by simulating actions for every $b \in B_t$, maintaining only the new belief points that are furthest away from all other points already in $B_{t+1}$. This scheme is a heuristic to let $B_t$ cover a wide area of the belief space, but comes at a cost as it requires computing distances between all $b \in B_t$. By backing up all $b \in B_t$ the PBVI algorithm generates at each stage approximately $|B_t|$ vectors, which can lead to slow performance in domains requiring large $B_t$.

In the next section we will present a point-based POMDP value iteration method which does not require backing up all $b \in B$. We compute backups for a subset of $B$ only, but seeing to it that the computed solution will be effective for the complete set $B$. As a result we limit the growth of the number of vectors in the successive value function estimates, leading to significant speedups.

## 3. Randomized Point-based Backup Stages

We have introduced the POMDP framework which models agents inhabiting stochastic environments that are partially observable to them, and discussed exact and approximate methods for computing successful plans for such agents. Below we describe Perseus, an approximate solution method capable of computing competitive solutions in large POMDP domains.

### 3.1 Perseus

Perseus is an approximate point-based value iteration algorithm for POMDPs (Vlassis & Spaan, 2004; Spaan & Vlassis, 2004). The value update scheme of Perseus implements a randomized approximate backup operator $\tilde{H}_{\text{Perseus}}$ that increases (or at least does not decrease) the value of all belief points in $B$. Such an operator can be very efficiently implemented in POMDPs given the shape of the value function. The key idea is that in each value backup stage we can improve the value of *all* points in the belief set by only updating the value and its gradient of a (randomly selected) subset of the points. In each backup stage, given a value function $V_n$, we compute a value function $V_{n+1}$ that improves the value of all $b \in B$, i.e., we build a value function $V_{n+1} = \tilde{H}_{\text{Perseus}} V_n$ that upper bounds $V_n$ over $B$ (but not necessarily over $\Delta$ which would require linear programming):

$$V_n(b) \leq V_{n+1}(b), \qquad \text{for all } b \in B. \tag{16}$$

We first let the agent randomly explore the environment and collect a set $B$ of reachable belief points, which remains fixed throughout the complete algorithm. We initialize the value function $V_0$ as a single vector with all its components equal to $\frac{1}{1-\gamma} \min_{s,a} r(s,a)$ (Zhang & Zhang, 2001). Starting with $V_0$, Perseus performs a number of backup stages until some convergence criterion is met. Each backup stage is defined as follows (where $\tilde{B}$ is an auxiliary set containing the non-improved points):

---

Perseus backup stage: $V_{n+1} = \tilde{H}_{\text{Perseus}} V_n$

1. Set $V_{n+1} = \emptyset$. Initialize $\tilde{B}$ to $B$.

2. Sample a belief point $b$ uniformly at random from $\tilde{B}$ and compute $\alpha = \texttt{backup}(b)$.

3. If $b \cdot \alpha \geq V_n(b)$ then add $\alpha$ to $V_{n+1}$, otherwise add $\alpha' = \arg\max_{\{\alpha_n^i\}_i} b \cdot \alpha_n^i$ to $V_{n+1}$.

4. Compute $\tilde{B} = \{b \in B : V_{n+1}(b) < V_n(b)\}$.

5. If $\tilde{B} = \emptyset$ then stop, else go to 2.

---

Often, a small number of vectors will be sufficient to improve $V_n(b)\ \forall b \in B$, especially in the first steps of value iteration. The idea is to compute these vectors in a randomized greedy manner by sampling from $\tilde{B}$, an increasingly smaller subset of $B$. We keep track of the set of non-improved points $\tilde{B}$ consisting of those $b \in B$ whose new value $V_{n+1}(b)$ is still lower than $V_n(b)$. At the start of each backup stage, $V_{n+1}$ is set to $\emptyset$ which means $\tilde{B}$ is initialized to $B$, indicating that all $b \in B$ still need to be improved in this backup stage. As long as $\tilde{B}$ is not empty, we sample a point $b$ from $\tilde{B}$ and compute $\alpha = \texttt{backup}(b)$. If $\alpha$ improves the value of $b$ (i.e., if $b \cdot \alpha \geq V_n(b)$ in step 3), we add $\alpha$ to $V_{n+1}$ and update $V_{n+1}(b)$ for all $b \in B$ by computing their inner product with the new $\alpha$. The hope is that $\alpha$ improves the value of many other points in $B$, and all these points are removed from $\tilde{B}$. As long as $\tilde{B}$ is not empty we sample belief points from it and add their $\alpha$ vectors.

To ensure termination of each backup stage we have to enforce that $\tilde{B}$ shrinks when adding vectors, i.e., that each $\alpha$ actually improves at least the value of the $b$ that generated it. If not (i.e., $b \cdot \alpha < V_n(b)$ in step 3), we ignore $\alpha$ and insert a copy of the maximizing vector of $b$ from $V_n$ in $V_{n+1}$. Point $b$ is now considered improved and is removed from $\tilde{B}$ in step 4, together with any other belief points which had the same vector as maximizing one in $V_n$. This procedure ensures that $\tilde{B}$ shrinks and the backup stage will terminate. A pictorial example of a backup stage is presented in Fig. 1.

Perseus performs backup stages until some convergence criterion is met. For point-based methods several convergence criteria can be considered, one could for instance bound the difference between successive value function estimates $\max_{b \in B}(V_{n+1}(b) - V_n(b))$. Another option would be to track the number of policy changes: the number of $b \in B$ which had a different optimal action in $V_n$ compared to $V_{n+1}$ (Lovejoy, 1991).

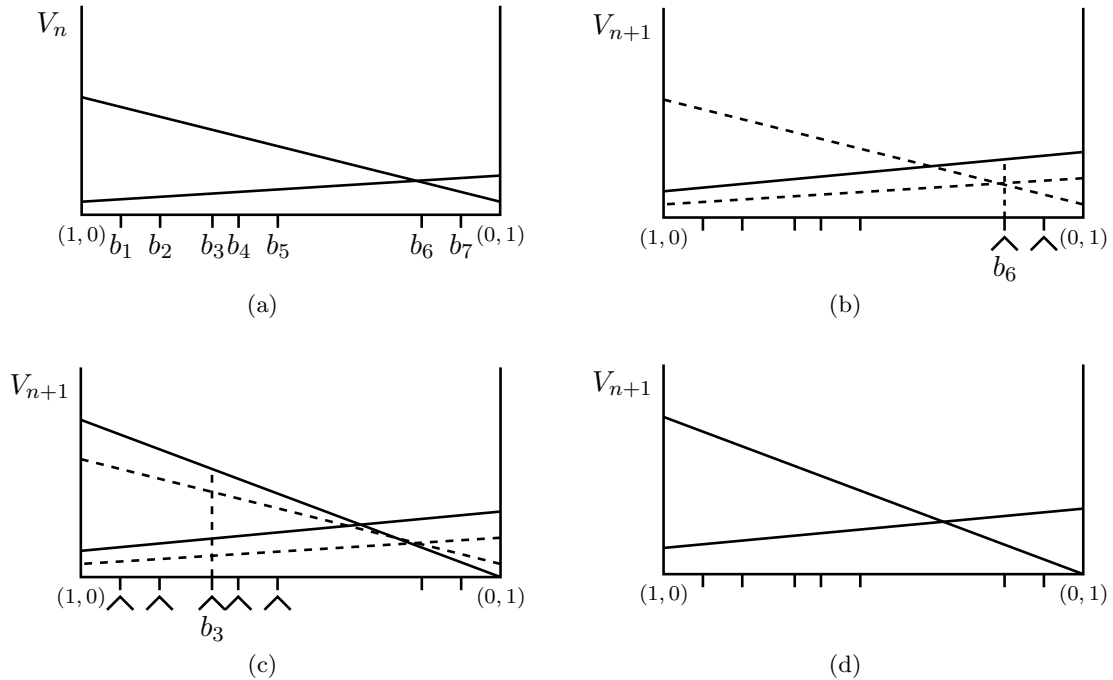

Figure 1: Example of a PERSEUS backup stage in a two state POMDP. The belief space is depicted on the $x$-axis and the $y$-axis represents $V(b)$. Solid lines are $\alpha_n^i$ vectors from the current stage $n$ and dashed lines are $\alpha_{n-1}^i$ vectors from the previous stage. We operate on a $B$ of 7 beliefs, indicated by the tick marks. The backup stage computing $V_{n+1}$ from $V_n$ proceeds as follows: (a) value function at stage $n$; (b) start computing $V_{n+1}$ by sampling $b_6$, add $\alpha = \texttt{backup}(b_6)$ to $V_{n+1}$ which improves the value of $b_6$ and $b_7$; (c) sample $b_3$ from $\{b_1, \ldots, b_5\}$, add $\texttt{backup}(b_3)$ to $V_{n+1}$ which improves $b_1$ through $b_5$; and (d) the value of all $b \in B$ has improved, the backup stage is finished.

### 3.2 Discussion

The key observation underlying the PERSEUS algorithm is that when a belief $b$ is backed up, the resulting vector improves not only $V(b)$ but often also the value of many other belief points in $B$. This results in value functions with a relatively small number of vectors (as compared to, e.g., Poon, 2001; Pineau et al., 2003). Experiments show indeed that the number of vectors grows modestly with the number of backup stages ($|V_n| \ll |B|$). In practice this means that we can afford to use a much larger $B$ than other point-based methods, which has a positive effect on the approximation accuracy as dictated by the bounds of Pineau et al. (2003). Furthermore, compared with other methods that build the set $B$ based on various heuristics (Pineau et al., 2003; Smith & Simmons, 2004), our build-up of $B$ is cheap as it only requires sampling random trajectories starting from $b_0$. Moreover, duplicate entries in $B$ will only affect the probability that a particular $b$ will be sampled in the value update stages, but not the size of $V_n$.

An alternative to using a single fixed set $B$ that is collected by following a fixed policy at the beginning of the algorithm, would be to resample a new $B_t$ after every $t$-th backup stage (or at fixed intervals) by following the most recent policy. Such an approach could be justified by the fact that an agent executing an optimal policy will most probably visit only a (small) subset of the beliefs in $B$. We have not tested how such a scheme would affect the solution quality of Perseus and what trade-offs we can achieve for the additional computational cost of sampling multiple sets $B$. We note that similar 'off-policy' learning using a fixed set of sampled states has also been adopted by other recent algorithms like LSPI (Lagoudakis & Parr, 2003) and PSDP (Bagnell, Kakade, Ng, & Schneider, 2004).

The backups of Perseus on a fixed set $B$ can be viewed as a particular instance of asynchronous dynamic programming (Bertsekas & Tsitsiklis, 1989). In asynchronous dynamic programming algorithms no full sweeps over the state space are made, but the order in which states are backed up is arbitrary. This allows an algorithm to focus on backups which may have a high potential impact, as for instance in the prioritized sweeping algorithm for solving fully observable MDPs (Moore & Atkeson, 1993; Andre, Friedman, & Parr, 1998). A drawback is that the notion of an exact planning horizon is somewhat lost: in general, after performing $n$ backup stages the computed plan will not be considering $n$ steps into the future, but less. By backing up non-improved belief points asynchronously Perseus focuses on interesting regions of the (reachable) belief space, and by sampling at random ensures that eventually all $b \in B$ will be taken into account. As we ensure that the value of a particular belief point never decreases, we are guaranteed that Perseus will converge: the proof only requires observing that every added vector is always below $V^*$ (Poon, 2001; Vlassis & Spaan, 2004). Moreover, as we explained above, Perseus can handle large belief sets $B$, thus obviating the use of dynamic belief point selection strategies like those proposed by Hauskrecht (2000), Poon (2001), and Pineau et al. (2003). Note that the only parameter to be set by the user is the size of $B$; however, the complexity of the resulting policy seems to be only mildly dependent on the size of $B$.

An interesting issue is how many new vectors are generated in each backup stage of Perseus, and how this may affect the speed of convergence of the algorithm. In general, the smaller the size $|V_n|$ of a value function, the faster the backups (since the `backup` operator has linear dependence on $|V_n|$). On the other hand, two consecutive value functions may differ arbitrarily in size—and we have observed cases where the new value function has fewer vectors than the old value function—which makes it hard to derive bounds on the speed of convergence of Perseus and complicates the analysis of the involved trade-offs. We have mainly identified two cases where only a small number of new vectors are added to a value function. The first case is during the initial backup stages, and when $V_0$ has been initialized very low (e.g., for large $\gamma$ and large negative immediate reward). In this case a single vector may improve all points, for a number of backup stages, until the value function has reached some sufficient level. The second case is near convergence, when the value function has almost converged in certain regions of the belief space. Sampling a belief point in such a region will result in a (near) copy of the old vector. Whereas the former case provides evidence that the value function has been initialized too low (and adding a single vector is an efficient way to 'correct' this), the latter case may be viewed as providing evidence for the convergence of Perseus.

### 3.3 Extension to Very Large or Continuous Action Spaces

An attractive feature of Perseus is that it can be naturally extended to very large or continuous action spaces, due to the 'improve–only' principle of its backup stage. Note that the backup operator in (11) involves a maximization over all actions in $A$. When the action space $A$ is finite and small, one can cache in advance the transition, observation, and reward models for all $a \in A$, and therefore achieve an optimized implementation of the backup operator. For very large or continuous action spaces, the full maximization over actions in (11) is clearly infeasible, and one has to resort to sampling-based techniques. The idea here is to replace the full maximization over actions with a *sampled max* operator that performs the maximization over a random subset of $A$ (Szepesvári & Littman, 1996). This also means that one has to compute the above models 'on the fly' for each sampled action, which requires an algorithm (a parameterized model family) that can generate all needed models for any action that is given as input. Such generated models can be cached for later use in case the same action is considered again in future iterations (see the experimental section for using these so-called 'old' actions).

The use of such a sampled max operator is very well suited for the backup scheme of Perseus in which we only require that the values of belief points do not decrease over two consecutive backup stages. In particular, we can replace the backup operator in (11) with a new backup operator $\alpha = \texttt{backup}'$ defined as follows (Spaan & Vlassis, 2005):

$$\texttt{backup}'(b) = \underset{\{g_a^b\}_{a \in A_b'}}{\arg\max}\, b \cdot g_a^b, \qquad (17)$$

where $A_b'$ is a random set of actions drawn from $A$, and $g_a^b$ is defined in (12). For each sampled action $a \in A_b'$ we generate the POMDP models on the fly as mentioned above, and from these models we compute the required vectors $g_a^b$ to be used in $\texttt{backup}'$.

The $\texttt{backup}'$ operator can now simply replace the backup operator in step 2 of Section 3.1. As in the full maximization case, we need to check in step 3 whether any of the vectors generated by the actions in $A_b'$ improves the value of the particular belief point. If not, we keep the old vector with its associated action that was selected in a previous backup stage. Concerning the sample complexity of the $\texttt{backup}'$ operator, we can derive simple bounds that involve the number of actions drawn and the probability to find a 'good' action from $A$ (good in terms of value improvement of $b$). We can easily show that with probability at least $1 - \delta$, the best action among $n = |A_b'|$ actions selected uniformly at random from $A$ is among the best $\epsilon$ fraction of all actions from $A$, if $n \geq \lceil \log \delta / \log(1-\epsilon) \rceil$.

In practice various sampling schemes are possible, which vary in the way $A_b'$ is constructed. We have identified a number of proposal distributions from which to sample actions: (1) uniform from $A$, (2) a Gaussian distribution centered on the best known action for the particular $b$, i.e., $a(\alpha_n^b)$, and (3) a Dirac distribution on $a(\alpha_n^b)$. The latter two take into account the policy computed so far by focusing on the current action associated with the input belief $b$ (as recorded in $V_n$), while sampling uniformly at random uses no such knowledge. Actions sampled uniformly at random can be viewed as exploring actions, while the other two distributions are exploiting current knowledge. As we can select the makeup of $A_b'$, we can choose any combination of the distributions mentioned above, allowing us to explore and exploit at the same time. In our experiments (see Section 5.2) we implement the $\texttt{backup}'$ operator using a number of different combinations and analyze their effects.

## 4. Related Work

In Section 2.2 we reported on a class of approximate solution techniques for POMDPs that focus on computing a value function approximation based on a fixed set of prototype belief points. Here we will broaden the picture to other approximate POMDP solution methods. A related overview is provided by Hauskrecht (2000).

A few heuristic control strategies have been proposed which rely on a solution of the underlying MDP. A popular technique is $Q_{\text{MDP}}$ (Littman, Cassandra, & Kaelbling, 1995), a simple approximation technique that treats the POMDP as if it were fully observable and solves the MDP, e.g., using value iteration. The resulting $Q(s, a)$ values are used to define a control policy by $\pi(b) = \arg\max_a \sum_s b(s) Q(s, a)$. $Q_{\text{MDP}}$ can be very effective in some domains, but the policies it computes will not take informative actions, as the $Q_{\text{MDP}}$ solution assumes that any uncertainty regarding the state will disappear after taking one action. As such, $Q_{\text{MDP}}$ policies will fail in domains where repeated information gathering is necessary.

One way to sidestep the intractability of exact POMDP value iteration is to grid the belief simplex, using either a fixed grid (Lovejoy, 1991; Bonet, 2002) or a variable grid (Brafman, 1997; Zhou & Hansen, 2001). Value backups are performed for every grid point, but only the value of each grid point is preserved and the gradient is ignored. The value of non-grid points is defined by an interpolation rule. The grid based methods differ mainly on how the grid points are selected and what shape the interpolation function takes. In general, regular grids do not scale well in problems with high dimensionality and non-regular grids suffer from expensive interpolation routines.

An alternative to computing an (approximate) value function is policy search: these methods search for a good policy within a restricted class of controllers. For instance, policy iteration (Hansen, 1998b) and bounded policy iteration (BPI) (Poupart & Boutilier, 2004) search through the space of (bounded-size) stochastic finite state controllers by performing policy iteration steps. Other options for searching the policy space include gradient ascent (Meuleau, Kim, Kaelbling, & Cassandra, 1999; Kearns, Mansour, & Ng, 2000; Ng & Jordan, 2000; Baxter & Bartlett, 2001; Aberdeen & Baxter, 2002) and heuristic methods like stochastic local search (Braziunas & Boutilier, 2004). In particular, the PEGASUS method (Ng & Jordan, 2000) estimates the value of a policy by simulating a (bounded) number of trajectories from the POMDP using a fixed random seed, and then takes steps in the policy space in order to maximize this value. Policy search methods have demonstrated success in several cases, but searching in the policy space can often be difficult and prone to local optima.

Another approach for solving POMDPs is based on heuristic search (Satia & Lave, 1973; Hansen, 1998a; Smith & Simmons, 2004). Defining an initial belief $b_0$ as the root node, these methods build a tree that branches over $(a, o)$ pairs, each of which recursively induces a new belief node. These methods bear a similarity to PERSEUS since they also focus on reachable beliefs from $b_0$. However, they differ in the way belief points are selected to back up; in the above methods branch and bound techniques are used to maintain upper and lower bounds to the expected return at fringe nodes in the search tree. Hansen (1998a) proposes a policy iteration method that represents a policy as a finite state controller, and which uses the belief tree to focus the search on areas of the belief space where the controller can most likely be improved. However, its applicability to large problems is limited by its use of full dynamic programming updates. HSVI (Smith & Simmons, 2004) is an approximate value iteration technique that performs a heuristic search through the belief space for beliefs at which to update the bounds, similar to work by Satia and Lave (1973). An alternative recent approach to maintaining uncertainty estimates of an approximate value function is based on Gaussian Processes (Tuttle & Ghahramani, 2004).

| Name | $|S|$ | $|O|$ | $|A|$ |
|---|---|---|---|
| Tiger-grid | 33 | 17 | 5 |
| Hallway | 57 | 21 | 5 |
| Hallway2 | 89 | 17 | 5 |
| Tag | 870 | 30 | 5 |
| Continuous navigation | 200 | 16 | $\infty$ |
| cTRC | 200 | 10 | $\infty$ |

Table 1: Characteristics of problem domains.

Compression techniques can be applied to large POMDPs to reduce the dimensionality of the belief space, facilitating the computation of an approximate solution. Roy, Gordon, and Thrun (2005) apply Exponential family PCA to a sample set of beliefs to find a low-dimensional representation, based on which an approximate solution is sought. Such a non-linear compression can be very effective, but requires learning a reward and transition model in the reduced space. After such a model is learned, one can compute an approximate solution for the original POMDP using, e.g., MDP value iteration. Alternatively linear compression techniques can be used which preserve the shape of value function (Poupart & Boutilier, 2003). Such a property is desirable as it allows one to exploit the existing POMDP machinery. For instance, linear compression has been applied as a preprocessing step for BPI (Poupart & Boutilier, 2005) as well as PERSEUS (Poupart, 2005).

The literature on POMDPs with continuous actions is still relatively sparse (Thrun, 2000; Ng & Jordan, 2000; Baxter & Bartlett, 2001). Thrun (2000) applies real-time dynamic programming on a POMDP with a continuous state and action space. In that work beliefs are represented by sets of samples drawn from the state space, while $Q(b, a)$ values are approximated by nearest-neighbor interpolation from a (growing) set of prototype values and are updated by on-line exploration and the use of sampling-based Bellman backups. PEGASUS can also handle continuous action spaces, at the cost of a sample complexity that is polynomial in the size of the state space (Theorem 3, Ng & Jordan, 2000).

## 5. Experiments

We will show experimental results applying PERSEUS on benchmark problems from the POMDP literature, and present two POMDP domains for testing PERSEUS in problems with continuous action spaces. Table 5 summarizes these domains in terms of the size of $S$, $O$ and $A$. Each belief set was gathered by simulating trajectories of interactions of the agent with the POMDP environment starting at a random state sampled from $b_0$, and at each time step the agent picked an action uniformly at random. In all domains the discount factor $\gamma$ was set to 0.95.

### 5.1 Discrete Action Spaces

The Hallway, Hallway2 and Tiger-grid problems (introduced by Littman et al., 1995) are maze domains that have been commonly used to test scalable POMDP solution techniques (Littman et al., 1995; Brafman, 1997; Zhou & Hansen, 2001; Pineau et al., 2003; Smith & Simmons, 2004; Spaan & Vlassis, 2004; Poupart, 2005). The Tag domain (Pineau et al., 2003) is an order of magnitude larger than the first three problems, and is a recent benchmark problem (Pineau et al., 2003; Smith & Simmons, 2004; Braziunas & Boutilier, 2004; Poupart & Boutilier, 2004; Spaan & Vlassis, 2004; Poupart, 2005).

#### 5.1.1 Benchmark Mazes

Littman et al. (1995) introduced three benchmark maze domains: Tiger-grid, Hallway, and Hallway2. All of them are navigation tasks: the objective for an agent is to reach a designated goal state as quickly as possible. The agent observes each possible combination of the presence of a wall in four directions plus a unique observation indicating the goal state; in the Hallway problem three other landmarks are also available. At each step the agent can take one out of five actions: {*stay in place, move forward, turn right, turn left, turn around*}. Both the transition and the observation model are noisy. Table 2(a) through (c) compares the performance of Perseus to other algorithms. For each problem we sampled a set $B$ of 1,000 beliefs, and executed Perseus 10 times for each problem using different random seeds. The average expected discounted reward R is computed from 1,000 trajectories starting from random states (drawn according to $b_0$) for each of the 10 Perseus runs, and following the computed policy. The reported reward $R$ is the average over these 10,000 trajectories. Perseus reaches competitive control quality using a small number of vectors resulting in a considerable speedup.[2]

#### 5.1.2 Tag

The goal in the Tag domain, described by Pineau et al. (2003), is for a robot to search for a moving opponent robot and tag it. The chasing robot cannot observe the opponent until they occupy the same position, at which time it should execute the *tag* action in order to win the game, and receive a reward of 10. If the opponent is not present at the same location, the reward will be $-10$, and the robot is penalized with a $-1$ reward for each motion action it takes. The opponent tries to escape from being tagged by moving away of the chasing robot, however, it has a 0.2 probability of remaining at its location. The chasing and opponent robot both start at a random location. The chasing robot has perfect information regarding its own position and its movement actions {*north, east, south, west*} are deterministic. The state space is represented as the cross-product of the states of the two robots. Both robots can be located in one of the 29 positions depicted in Fig. 2(a), and the opponent can also be in a *tagged* state, resulting in a total of 870 states. Tag is a rather large benchmark problem compared to other POMDP problems studied in literature, but it exhibits a sparse structure. We applied Perseus to a belief set $B$ of 10,000 points.

---

2. Perseus and $Q_{\text{MDP}}$ results (in Section 5.1) were computed in Matlab on an Intel Pentium IV 2.4 GHz; other results were obtained on different platforms, so time comparisons are rough.

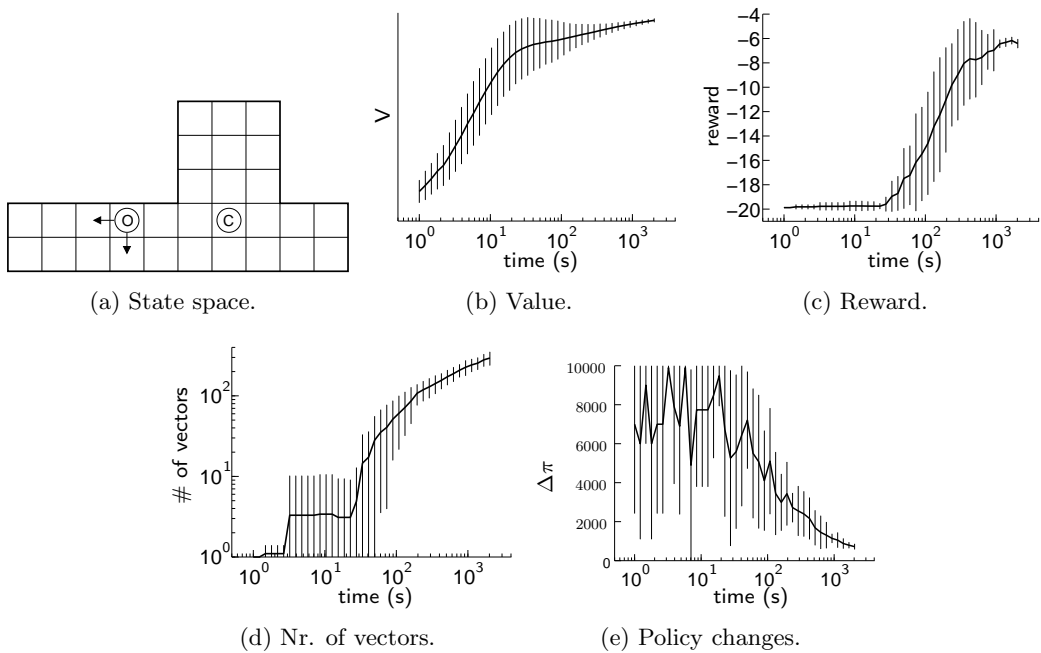

Figure 2: Tag: (a) state space with chasing and opponent robot; (b)–(e) performance of Perseus.

In Fig. 2(b)–(e) we show the performance of Perseus averaged over 10 runs, where error bars indicate standard deviation within these runs. To evaluate the computed policies we tested each of them on 10 trajectories (of at most 100 steps) times 100 starting positions (sampled from the starting belief $b_0$). Fig. 2(b) displays the value as estimated on $B$, $\sum_{b \in B} V(b)$; (c) the expected discounted reward averaged over the 1,000 trajectories; (d) the number of vectors in the value function estimate, $|\{\alpha_n^i\}|$; and (e) the number of policy changes: the number of $b \in B$ which had a different optimal action in $V_{n-1}$ compared to $V_n$. The latter can be regarded as a measure of convergence for point-based solution methods (Lovejoy, 1991). We can see that in almost all experiments Perseus reaches solutions of virtually equal quality and size.

Table 2(d) compares the performance of Perseus with other state-of-the-art methods. The results show that in the Tag problem Perseus displays better control quality than any other method and computes its solution an order of magnitude faster than most other methods. Specifically, its solution computed on $|B| = 10,000$ beliefs consists of only 280 vectors, much less than PBVI which maintains a vector for each of its 1334 $b \in B$. This indicates that the randomized backup stage of Perseus is justified: it takes advantage of a large $B$ while the size of the value function grows moderately with the planning horizon, leading to significant speedups. It is interesting to compare the two variations of BPI, with bias (w/b) (Poupart, 2005) or without (n/b) (Poupart & Boutilier, 2004). The bias focuses

| **Tiger-grid** | R | $|\pi|$ | T |
|---:|---|---|---|
| HSVI | 2.35 | 4860 | 10341 |
| Perseus | 2.34 | 134 | 104 |
| PBUA | 2.30 | 660 | 12116 |
| PBVI | 2.25 | 470 | 3448 |
| BPI w/b | 2.22 | 120 | 1000 |
| Grid | 0.94 | 174 | n.a. |
| $Q_{\mathrm{MDP}}$ | 0.23 | n.a. | 2.76 |

(a) Results for Tiger-grid.

| **Hallway** | R | $|\pi|$ | T |
|---:|---|---|---|
| PBVI | 0.53 | 86 | 288 |
| PBUA | 0.53 | 300 | 450 |
| HSVI | 0.52 | 1341 | 10836 |
| Perseus | 0.51 | 55 | 35 |
| BPI w/b | 0.51 | 43 | 185 |
| $Q_{\mathrm{MDP}}$ | 0.27 | n.a. | 1.34 |

(b) Results for Hallway.

| **Hallway2** | R | $|\pi|$ | T |
|---:|---|---|---|
| Perseus | 0.35 | 56 | 10 |
| HSVI | 0.35 | 1571 | 10010 |
| PBUA | 0.35 | 1840 | 27898 |
| PBVI | 0.34 | 95 | 360 |
| BPI w/b | 0.32 | 60 | 790 |
| $Q_{\mathrm{MDP}}$ | 0.09 | n.a. | 2.23 |

(c) Results for Hallway2.

| **Tag** | R | $|\pi|$ | T |
|---:|---|---|---|
| Perseus | $-6.17$ | 280 | 1670 |
| HSVI | $-6.37$ | 1657 | 10113 |
| BPI w/b | $-6.65$ | 17 | 250 |
| BBSLS | $\approx -8.3$ | 30 | $10^5$ |
| BPI n/b | $-9.18$ | 940 | 59772 |
| PBVI | $-9.18$ | 1334 | 180880 |
| $Q_{\mathrm{MDP}}$ | $-16.9$ | n.a. | 16.1 |

(d) Results for Tag.

Table 2: Experimental comparisons of Perseus with other algorithms. Perseus results are averaged over 10 runs. Each table lists the method, the average expected discounted reward R, the size of the solution $|\pi|$ (value function or controller size), and the time T (in seconds) used to compute the solution. Sources: PBVI (Pineau et al., 2003), BPI no bias (Poupart & Boutilier, 2004), BPI with bias (Poupart, 2005), HSVI (Smith & Simmons, 2004), Grid (Brafman, 1997), PBUA (Poon, 2001), and BBSLS (Braziunas & Boutilier, 2004) (approximate, read from figure).

on the reachable belief space by incorporating the initial belief which dramatically increases its performance in solution size and computation time, but it does not reach the control quality of Perseus.

### 5.2 Continuous Action Spaces

We applied Perseus in two domains with continuous action spaces: an agent equipped with proximity sensors moving at a continuous heading and distance, and a navigation task involving a mobile robot with omnidirectional vision in a perceptually aliased office environment.

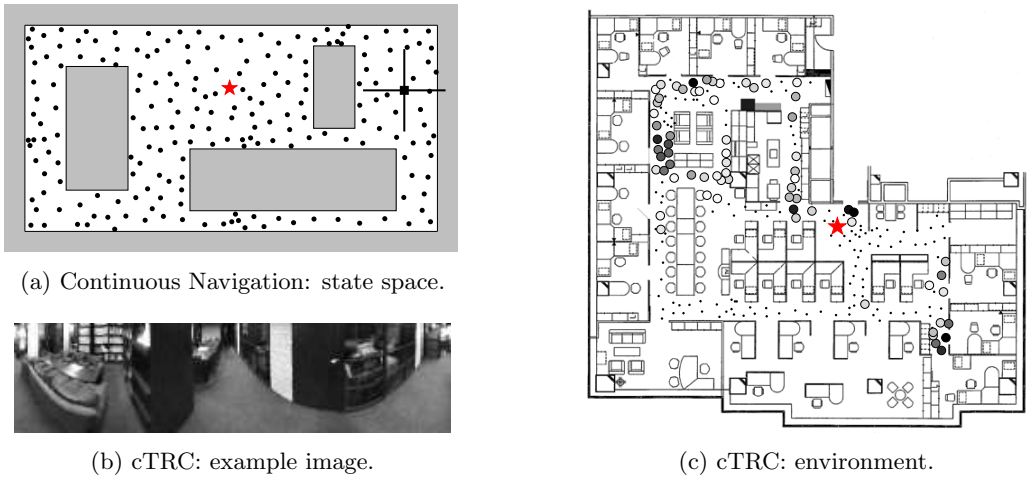

(a) Continuous Navigation: state space.

(b) cTRC: example image.

(c) cTRC: environment.

Figure 3: Continuous action space domains: the points indicate the states, ★ depicts the goal state. (a) Environment of the Continuous Navigation problem: the black square represents the agent, the four beams indicate the range of its proximity sensors. (b) cTRC Problem: panoramic image corresponding to a prototype feature vector $o_k \in O$, and (c) its induced $p(s|o_k)$. The darker the dot, the higher the probability.

5.2.1 Continuous Navigation

We first tested our approach on a navigation task in a simulated environment, in which an agent can move at a continuous heading and distance. The Continuous Navigation environment represents a $20 \times 10$m hallway which is highly perceptually aliased (see Fig. 3(a)). The agent inhabiting the hallway is equipped with four proximity sensors, each observing one compass direction. We assume that a proximity sensor can only detect whether there is a wall within its range of 2m or not, resulting in a total number of 16 possible sensor readings. The agent's sensor system is noisy: with 0.9 probability the correct wall configuration is observed, otherwise one of the other 15 observations is returned with equal probability. The task is to reach a goal location located in an open area where there are no walls near enough for the agent to detect. The agent is initialized at a random state in the environment, and it should learn what movement actions to take in order to reach the goal as fast as possible.

As Perseus assumes a finite and discrete state space $S$ (the set of all possible locations of the agent) we need to discretize this space; we performed a simple $k$-means clustering on a random subset of all possible locations, resulting in a grid of 200 locations depicted in Fig. 3(a). The agent's actions are defined by two parameters: the heading $\theta$ to which the agent turns and the distance $d$ it intends to move in this direction. Executing an action transports it according to a Gaussian distribution centered on the expected resulting position, which is defined as its current $(x, y)$ position translated $d$ meter in the direction $\theta$. The standard deviation of the Gaussian transition model is $0.25d\,\mathbf{I}$, which means the further the agent wants to travel, the more uncertainty there will be regarding its resulting position. The distance parameter $d$ is limited to the interval $[0,2]$m and the heading $\theta$ ranges on $[0, 2\pi]$. Each movement is penalized with a reward of $-0.1$ per step and the reward obtainable at the goal location is 10.

To test the feasibility of PERSEUS in continuous action spaces, i.e., whether it can compute successful policies by sampling actions at random, we experimented with a number of different sampling schemes for the `backup'` operator. Each scheme is defined by the makeup of $A'_b = \{A^{\mathcal{U}}, A^{\mathcal{N}}_b, A^{old}_b\}$, which is composed of samples from three distributions: $A^{\mathcal{U}}$: uniformly at random; $A^{\mathcal{N}}_b$: a Gaussian distribution centered on the best known action $a(\alpha^b_n)$ for $b$ so far, with standard deviation $\sigma_\theta = \frac{\pi}{5}$ for $\theta$ and $\sigma_d = 0.1$ for $d$; and $A^{old}_b$: a Dirac distribution on the best known action. We will describe $A'_b$ by the number of samples from each distribution $\{|A^{\mathcal{U}}|, |A^{\mathcal{N}}_b|, |A^{old}_b|\}$. We tested the following schemes: sampling a single action uniformly at random $\{1,0,0\}$, or from a Gaussian distribution on $a(\alpha^b_n)$ $\{0,1,0\}$; adding $a(\alpha^b_n)$ to both schemes resulting in $\{1,0,1\}$ and $\{0,1,1\}$; and $\{k,k,1\}$, sampling $k$ actions from the uniform and Gaussian distributions and including the old action. The latter scheme explores the option of sampling more than one action from a particular distribution, and we tested $k = \{1, 3, 10\}$. The option to try the best known ('old') action for the particular $b$ is relatively cheap as we can cache its transition, observation, and reward model the first time it is chosen (at a previous backup stage).

In this problem we used a set $B$ of 10,000 belief points. To evaluate the control quality of the computed value functions we collected rewards by sampling 10 trajectories from 100 random starting locations at particular time intervals, while following the policy computed so far. Each trajectory was stopped after a maximum of 100 steps (if the agent had not reached the goal by then), and the collected reward was properly discounted. All results are averaged over 10 runs of PERSEUS with a different random seed and are computed in Matlab on an Intel Xeon 3.4GHz.

Fig. 4 shows the results for each of the sampling schemes mentioned above. The top row displays the control quality as indicated by the average discounted reward. In Fig. 4(a) we can see that just sampling a single action uniformly at random $\{1,0,0\}$ already gives good performance, while extending $A'_b$ to include the best known action $\{1,0,1\}$ improves control quality. The Gaussian sampling schemes $\{0,1,0\}$ and $\{0,1,1\}$ learn slower as they can take only small steps in action space. An additional disadvantage of Gaussian sampling is the need for the user to specify the standard deviation. Fig. 4(b) depicts the control quality of the schemes in which we sample from three distributions $\{k,k,1\}$, for different values of $k$. The figure shows that all tested variations reach similar control quality, but trying more actions for a particular $b$ can slow down learning. However, when looking at the size of the value function (Fig. 4(c)–(d)), we see that for $k = 10$ the resulting value function is smaller than for any other scheme tested. It appears in this experiment that sampling more actions increases the chance of finding a high quality action that generalizes well (so fewer vectors are eventually needed to reach the same control quality), but at a higher computational cost per backup stage. Note that for all tested schemes the number of vectors in the value function remains two orders of magnitude lower than the size of $B$ (10,000 belief points), confirming the efficient behavior of the PERSEUS randomized backup scheme.

To obtain more insight in the effect of sampling from different distributions in $A'_b$, we computed the relative frequency of occurrence of the maximizing action in $A^{\mathcal{U}}$, $A^{\mathcal{N}}_b$, and $A^{old}_b$. When executing a `backup'` we check whether the vector computed using the returned action actually improves $V(b)$, and if so, we record whether this action originated from $A^{\mathcal{U}}$, $A_b^{\mathcal{N}}$, or $A_b^{old}$. For every backup stage we normalize these counts with respect to the total number of backups in that backup stage (including those that did not improve $V(b)$). The resulting frequencies are plotted on the bottom row of Fig. 4 for two sampling schemes: sampling uniform and old $\{1, 0, 1\}$ (Fig. 4(e)) and sampling one action from all three distributions $\{1, 1, 1\}$ (Fig. 4(f)). We can see that over time the relative frequency of the best known action grows ("Improved: Old"), while the number of instances in which none of the sampled actions improves $V(b)$ drops to almost zero ("Not improved"). The frequencies of actions sampled from an uniform or Gaussian distribution ("Improved: Uniform" resp. "Improved: Gauss") resulting in the best action in $A_b'$ (and improving $V(b)$) also drop. These observations confirm the intuition that by sampling actions at random Perseus can effectively explore the action space (which is advantageous at the beginning of the algorithm), while as time progresses the algorithm seems to be able to exploit the actions that turn out to be useful.

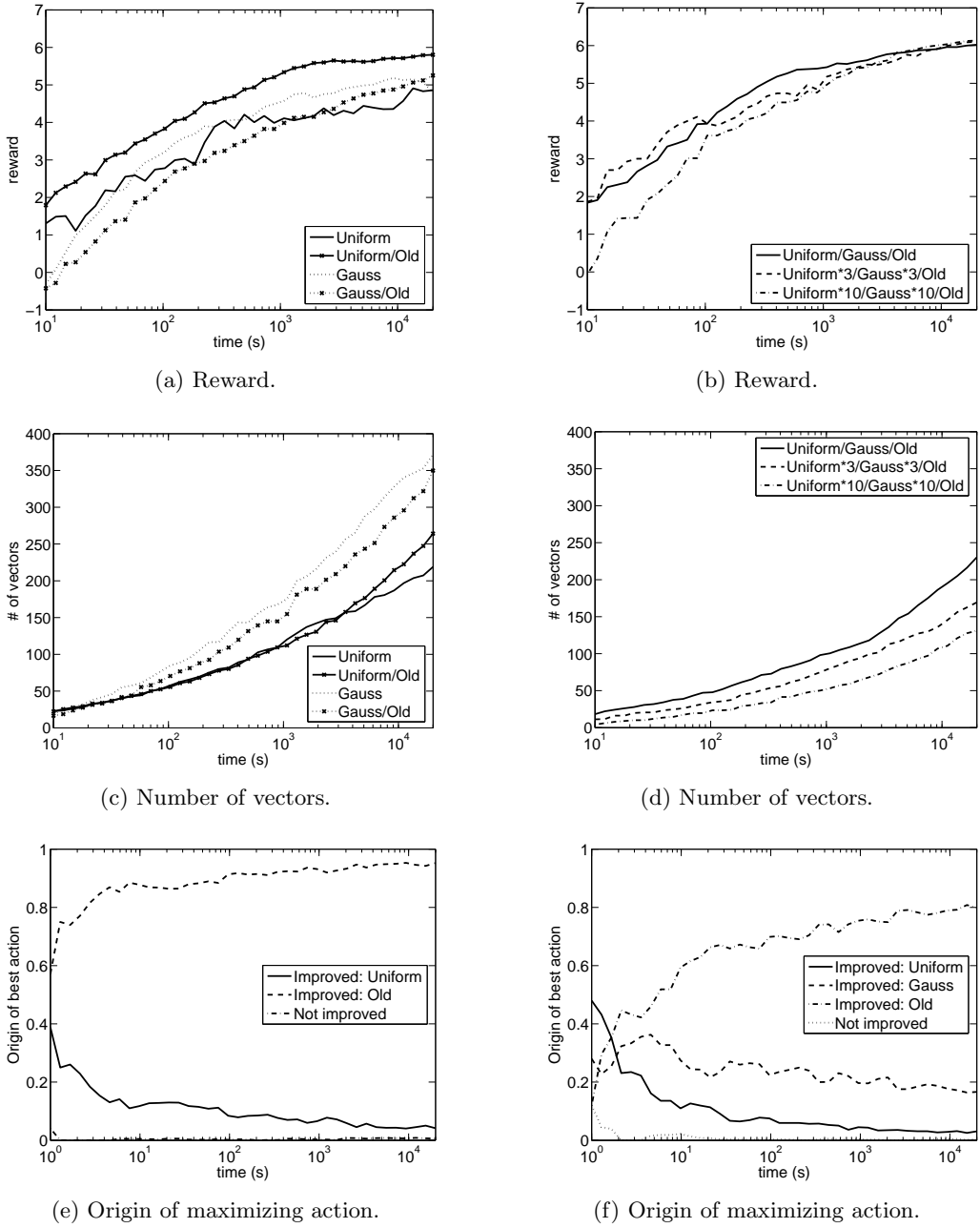

Figure 4: Perseus results on the Continuous Navigation problem, averaged over 10 runs. The left column shows the performance of $\{|A^{\mathcal{U}}|, |A_b^{\mathcal{N}}|, |A_b^{old}|\} = \{\{1,0,0\}, \{1,0,1\}, \{0,1,0\}, \{0,1,1\}\}$, and the right column displays $\{k, k, 1\}$ for $k = \{1, 3, 10\}$. The top row figures display the average discounted reward obtained vs. computation time, the figures in the middle row show the size of the value function, and the bottom row details the origin of the maximizing vector (see main text).

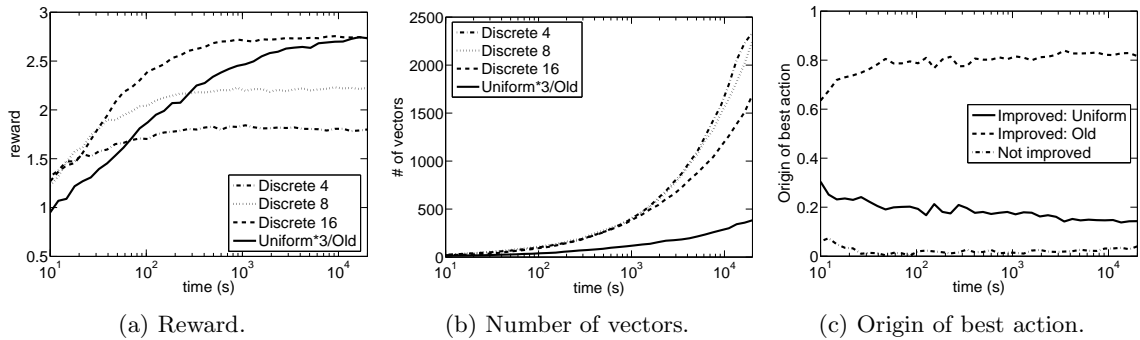

Figure 5: Performance of Perseus in cTRC domain, averaged over 10 runs.

### 5.2.2 Arbitrary Heading Navigation

To evaluate Perseus with continuous actions on a more realistic problem and compare with discretized action spaces we also include the cTRC domain. In this problem (adapted from Spaan & Vlassis, 2005) a mobile robot with omnidirectional vision has to navigate in a highly perceptually aliased office environment (see Fig. 3(b) and (c)). We use the MEM-ORABLE[3] robot database that contains a set of approximately 8000 panoramic images collected manually by driving the robot around in a $17 \times 17$ meters office environment. The robot can decide to move 5m in an arbitrary direction, i.e., its actions are parameterized by its heading ranging on $[0, 2\pi]$. We applied the same technique as in the Continuous Navigation domain to grid our state space in 200 states (Fig. 3(c)) and assume a Gaussian error on the resulting position. For our observation model we compressed the images with PCA and applied $k$-means clustering to create 10 three-dimensional prototype feature vectors $\{o_1, \ldots, o_{10}\}$. Fig. 3(c) shows the inverse observation model $p(s|o)$ for one observation, and Fig. 3(b) displays the image in the database closest to this particular prototype obser-vation. The task is to reach a certain goal state at which a reward of 10 can be obtained; each action yields a reward of $-0.1$. The belief set $B$ contained 10,000 belief points.

---

3. The MEMORABLE database has been provided by the Tsukuba Research Center in Japan, for the Real World Computing project.

We compared the continuous action extension of Perseus to three discretized versions of this problem, in which we applied regular Perseus to a fixed discrete action set of 4, 8 or 16 headings with equal separation (offset with a random angle to prevent any bias). Fig. 5 displays results for Perseus with $\{|A^{\mathcal{U}}|, |A_b^{\mathcal{N}}|, |A_b^{old}|\} = \{3, 0, 1\}$ (other schemes turned out to give similar results), and the three discrete action spaces. Fig. 5(a) shows that sampling from a continuous $A$ results in the same control quality as in the discrete 16 version, but it needs more time to reach it (as the `backup'` requires to generate transition, observation and reward models on the fly). As the discrete cases benefit from an optimized implementation (we can cache all transition, observation and reward models) the continuous action scheme needs some computation time to match performance or outperform them. However, when employing the continuous scheme, Perseus exploits the ability to move at an arbitrary heading to find a better policy than the discrete 4 and 8 cases. We see that providing the robot with a more fine-grained action space leads to better control quality, and in this problem a discretization of 16 headings appears to be fine-grained enough for good control performance. Fig. 5(b) plots the number of vectors in the value function for each scheme, where we see that for reaching the same control quality the continuous and discrete 16 version need a similar amount of vectors. Fig. 5(c) shows the relative frequency of occurrence of the maximizing action in $A^{\mathcal{U}}$ or $A_b^{old}$, as detailed in Section 5.2.1. As in Fig. 4(e)–(f) we see that over time the best known action is exploited, while the frequency of instances in which no sampled action improves the value of $b$ is diminished to near zero.

## 6. Conclusions

The partially observable Markov decision process (POMDP) framework provides an attractive and principled model for sequential decision making under uncertainty. It models the interaction between an agent and the stochastic environment it inhabits. A POMDP assumes that the agent has imperfect information: parts of the environment are hidden from the agent's sensors. The goal is to compute a plan that allows the agent to act optimally given uncertainty in sensory input and the uncertain effect of executing an action. Unfortunately, the expressiveness of POMDPs is counterbalanced by the intractability of computing exact solutions, which calls for efficient approximate solution techniques. In this work we considered a recent line of research on approximate point-based POMDP algorithms that plan on a sampled set of belief points.

We presented Perseus, a randomized point-based value iteration algorithm for planning in POMDPs. Perseus operates on a large belief set sampled by simulating random trajectories through belief space. Approximate value iteration is performed on this belief set by applying a number of backup stages, ensuring that in each backup stage the value of each point in the belief set is improved; the key observation is that a single backup may improve the value of many belief points. Contrary to other point-based methods, Perseus backs up only a (randomly selected) subset of points in the belief set, sufficient for improving the value of each belief point in the set. Experiments confirm that this allows us to compute value functions that consist of only a small number of vectors (relative to the belief set size), leading to significant speedups. We performed experiments in benchmark problems

216

from literature, and Perseus turns out to be very competitive to other methods in terms of solution quality and computation time. We extended Perseus to compute plans for agents which have a very large or continuous set of actions at their disposal, by sampling actions from the action space. We demonstrated the viability of Perseus on two POMDP problems with continuous action spaces: a continuous navigation task and a robotic problem involving a mobile robot with omnidirectional vision. We analyzed a number of different action sampling schemes and compared with discretized action spaces.

Perseus has been recently extended to deal with structured state spaces (Poupart, 2005; Boger, Poupart, Hoey, Boutilier, Fernie, & Mihailidis, 2005), continuous observation spaces (Hoey & Poupart, 2005), and continuous state spaces (Porta, Spaan, & Vlassis, 2005). As future work we would like to explore alternative compact representations (Guestrin, Koller, & Parr, 2001; Theocharous, Murphy, & Kaelbling, 2004), as well as applying Perseus to cooperative multiagent domains, extending recent approaches (Emery-Montemerlo, Gordon, Schneider, & Thrun, 2004; Becker, Zilberstein, Lesser, & Goldman, 2004; Paquet, Tobin, & Chaib-draa, 2005).

## Acknowledgments

We would like to thank Bruno Scherrer, Geoff Gordon, Pascal Poupart, and the anonymous reviewers for their comments. This research is supported by PROGRESS, the embedded systems research program of the Dutch organization for Scientific Research NWO, the Dutch Ministry of Economic Affairs and the Technology Foundation STW, project AES 5414.